\documentclass[11pt]{article}
\usepackage[T1]{fontenc}
\usepackage[utf8]{inputenc}
\usepackage{lmodern}
\usepackage{microtype}
\usepackage[margin=1in]{geometry}
\usepackage{hyperref}

\title{Omnia: Synthetic Data Pipelines for Adaptive, Mission-Ready Militarized Humanoids\\
\large Transforming first-person spatial data from smart glasses into synthetic training pipelines that accelerate humanoid autonomy and resilience.}
\author{Mohammed Ayman Habib\\
\texttt{ayman@theomnia.io} \and Aldo Petruzzelli\\
\texttt{aldo@theomnia.io}}
\date{}

\begin{document}
\maketitle

 I. INTRODUCTION
At Omnia, we are developing a synthetic data--driven humanoid capability to accelerate both the baseline performance of militarized humanoids and the integration of advanced subsystems. Our core innovation is rapidly generating and refining high-fidelity training data from point-of-view (PoV) recordings, augmented reality headsets, and spatial browsing platforms. This pipeline produces scalable, domainspecific datasets, enabling humanoid perception, navigation, and decision-making systems to train in complex and contested environments---without the cost, risk, or time delays of physical field trials. Our synthetic data generation technology exposes humanoids to millions of simulated operational scenarios---from urban patrols to CBRNE reconnaissance---long before prototypes reach the field. This approach not only shortens development cycles, but also produces humanoids that are more adaptive, resilient, and mission-ready for dynamic combat environments. Omnia's platform is at TRL 5--6 and IRL 3--4. We have demonstrated synthetic environment creation, automated labeling, and perception-model training for humanoid-relevant tasks. Over the next 18--24 months, we will advance toward TRL 7--8 and IRL 7--8, scaling integration with humanoid prototypes and demonstrating end-to-end synthetic training pipelines that transfer directly to field experimentation. Our data-driven models enhance core humanoid functions---such as navigation, patrol, reconnaissance, and load transport---by training humanoids on terrain variation, formation behaviors, and small unit tactics in synthetic environments. We also apply synthetic data to accelerate the development of advanced subsystems, including AI autonomy (HAW 008-A/B), multi-modal perception (EO/IR, acoustic, and CBRNE sniffers, as per HAW 003/005-A), and counterdetection survivability (HAW 007-A/B). By generating adversarial datasets that mimic sensor deception and concealment, we prepare humanoids to withstand and adapt to real-world denial environments. Omnia's approach directly advances Humanoid Augmented Warfare by making humanoids: 
\begin{itemize}
\item  Trainable at scale: Synthetic data enables safe, repeatable exposure to rare, dangerous, and edge-case scenarios
\item  Rapidly adaptable: Datasets can be tuned to new geographies, missions, or threats in days rather than months
\item  Force-multiplying partners: humanoids become reliable squad members, capable of mission execution alongside human war fighters with reduced training and higher survivability By combining synthetic data generation with humanoid development, Omnia ensures the Department of War can field militarized humanoid systems faster, safer, and with greater mission relevance, addressing both the baseline expectation and advanced subsystem innovations.
II. MILITARY BENEFITS
A. Alignment with Baseline Priorities
Omnia's innovation directly aligns with the Army's priorities for militarized humanoid capabilities by providing a scalable method to train, validate, and accelerate deployment of humanoids that augment human warfighters in high-threat environments. 
\item  Force augmentation \& risk reduction: By generating synthetic training data across operational scenarios (urban patrols, obstacle reconnaissance, CBRNE clearing), our approach prepares humanoids to perform Warrior Skill Level 1 tasks while keeping human Soldiers out of immediate danger.
\item  Mission endurance \& survivability: Synthetic data trains humanoids to recognize and adapt to complex terrain, illumination, and weather conditions that soldiers encounter, ensuring consistent performance across patrol, checkpoint security, and sustainment tasks.
\item  Rapid adaptability: Unlike traditional training, synthetic datasets can be created for new mission geographies or threats within days, aligning with the Army's need to quickly scale humanoid deployment across theaters.
B. Alignment with Advanced Subsystem Priorities
\end{itemize}

\begin{itemize}
\item  HAW 008-A/B: AI-enabled autonomy and mission command: Our synthetic environments provide the missionrich datasets needed to train humanoids to not just follow orders but plan, adapt, and execute tasks autonomously.
\item  HAW 003/005-A: Advanced perception and information acquisition: Synthetic multimodal data (visual EO/IR, acoustic, CBRNE proxies) accelerates training of robust perception systems that enable reconnaissance, target acquisition, and early warning in denied environments.
\item  HAW 007-A/B: Counter-detection survivability: By simulating adversarial sensor systems in synthetic training loops, humanoids learn to operate with reduced signature, directly enhancing survivability. By equipping humanoids with robust AI models trained in millions of synthetic combat-relevant scenarios, Omnia ensures that they enter the field more resilient to surprise conditions, interoperable with soldiers and capable of extending operational reach without increasing soldier risk.
C. Advantage of Solution
Given limited resources and competing priorities, Army customers will select Omnia because we uniquely provide the force multiplier for all humanoid development efforts: scalable, mission-specific synthetic data. This ensures their investment in humanoids delivers operationally effective systems on the required timeline, outperforming both direct competitors and indirect substitutes.
D. Advancing the State of the Art
Omnia advances the state of the art in militarized humanoids by shifting from incremental hardware improvements to a
synthetic data--driven paradigm that enables scalable, rapid, and adaptive training of autonomy. Our technology generates millions of mission-relevant scenarios, urban patrols, CBRNE reconnaissance, electronic warfare, and contested environments that would be impossible or prohibitively costly to replicate physically. This allows humanoids to achieve AIenabled autonomy (HAW 008-A/B), robust perception, and counter-detection survivability (HAW 007-A/B) long before field deployment. The result is a high-impact leap forward: humanoids that are more adaptive, resilient, and mission-ready, directly advancing the Army's Humanoid Augmented Warfare vision.
III. TECHNICAL APPROACH
Omnia's innovation is built on sound, proven principles of computer vision, robotics, and AI model training, reinforced by established engineering practices in simulation and synthetic data generation. The foundation of our approach is the physics-based replication of operational environments combined with machine learning algorithms that improve through exposure to diverse, high-fidelity datasets. This mirrors established practices in autonomous driving, aerospace simulation, and defense training, where synthetic data has already accelerated capability maturation and reduced cost.Our pipeline is scientifically and technically feasible because: 
\item  Validated Methods: Synthetic data has been shown to close the performance gap in perception and autonomy across industries; Omnia adapts these same methods to militarized humanoid requirements.
\item  Physics-Based Fidelity: We model real-world sensor modalities (EO/IR, acoustic, CBRNE proxies) with underlying physics engines, ensuring synthetic data aligns with battlefield conditions and meets HAW 003, 005, and 006 requirements.
\item  Transfer to Hardware: Trained models are tested in both simulated and prototype humanoid platforms, demonstrating that behaviors learned in synthetic environments transfer effectively to field operations, aligning with HAW 002 mobility and HAW 008 behaviors.
\item  Scalable Engineering Integration: Our architecture is hardware-agnostic, enabling rapid integration with baseline humanoid systems and advanced subsystems, supporting interoperability and modularity as outlined in the Army's expectations (HAW 001--014).
\item  Risk Reduction: By addressing edge cases and dangerous scenarios in simulation first, we reduce risks to Soldiers, prototypes, and resources, supporting the Army's goal of safer, faster fielding of humanoid capabilities. In sum, Omnia's approach is not speculative, it is rooted in established scientific methods, proven engineering frameworks, and direct alignment to the Army's baseline and advanced humanoid capability requirements. This ensures feasibility and positions our innovation for successful transition into Humanoid Augmented Warfare operations.
TABLE I
ENABLING TECHNOLOGIES, RISKS, AND MITIGATIONS FOR MILITARIZED HUMANOID OBJECTIVES Enabling Technology Risk to Militarized Humanoid Objectives Mitigation / Resolution High-fidelity synthetic environment generation (physics-based rendering of EO/IR, acoustic, and CBRNE data) Risk of domain gap between synthetic and real-world conditions could reduce transferability of trained models Use hybrid training (synthetic + limited real data), physics-informed models, and Army-provided operational datasets to calibrate fidelity. Continuous field-validation loops reduce domain shift AI model training for autonomy and mission behaviors (HAW 008-A/B) Risk of overfitting to synthetic scenarios, leading to brittle decision-making in unanticipated conditions Leverage adversarial data generation to simulate edge cases; employ reinforcement learning with randomized environments to ensure generalization and robustness Integration with humanoid hardware platforms (baseline locomotion, manipulation, power systems) Risk of integration delays if hardware and software timelines are not synchronized Design pipeline to be hardware-agnostic, using modular APIs and simulation-in-the-loop testing. Partner with hardware primes to align integration milestones Counter-detection and survivability testing (HAW 007-A/B) Risk of underrepresenting advanced adversary sensor systems in synthetic environments Incorporate open-source and government-furnished intelligence on adversary EO/IR, radar, and EW systems to model realistic threats. Iteratively validate synthetic deception strategies against lab hardware-in-the-loop tests Long-duration power and endurance modeling (HAW 011) Risk that synthetic data cannot fully capture long-term degradation effects (e.g., thermal, wear, battery decay) Use synthetic environments for operational planning and training, while incorporating real-world hardware endurance testing for final calibration. Maintain modular architecture to update models as new endurance data is collected
A. Literature Review
Training only on randomized synthetic RGB images produced a real-world object detector accurate to around 1.5cm, robust to clutter and occlusion. The model directly controlled a physical robot to grasp in clutter [1]. Thus showing us
that synthetic to real world transfer can occur with high precision. Synthetic data reduces real labeling needs while improving accuracy. In semantic segmentation, models trained with game-rendered data plus only one-third of the CamVid real set outperformed models trained on the entire CamVid set---clear evidence that synthetic data cuts annotation cost without sacrificing performance [2]. Domain randomization coupled with small real fine-tuning beats real data only. For object detection on KITTI, networks pretrained on synthetic with domain randomization and then lightly fine-tuned on real data exceeded models trained on real data alone thus showing a practical recipe for reliable sim-to-real perception [3]. Deep RL controllers trained entirely in a physics simulator successfully executed agile gaits (trotting, galloping) on real quadruped robots after transfer---demonstrating that control behaviors learned in synthetic environments can survive realworld dynamics [4]. The Army is already operationalizing synthetic training. The FY2024 DOT\&E report documents an Army STE-LTS operational demonstration (Feb 2024) to support rapid prototyping to rapid fielding in 1QFY25, with findings that STE-LTS devices can improve lethality and survivability via more realistic force-on-force training (with reliability/integration fixes underway). This validates DoD demand for synthetic environments at scale [5]. Scale matters, synthetic unlocks orders-of-magnitude data. Widely used datasets like SynthText provide 800k images / 8 million labeled word instances, illustrating how synthetic pipelines deliver annotated volume that is infeasible to acquire and label in the real world on program timelines [6]. These results de-risk HAW 003/005 (perception \& information acquisition) and HAW 008-A/B (AI-enabled command \& autonomy) by showing that (1) perception models trained on synthetic generalize to real sensors, and (2) control policies learned in simulation can execute on hardware [7]. DoD's ongoing STE adoption demonstrates institutional acceptance of synthetic environments for readiness, survivability, and accelerated fielding, aligning with Humanoid Augmented Warfare goals to train/validate complex behaviors before contact.
B. Technical Team
Amy Seo brings a strong background in synthetic biology, biomaterials, and AI-enabled modeling, with a proven track record of integrating biological experimentation with computational intelligence. At the Korea Institute of Science and Technology, she engineered bacterial strains for cancer immunotherapy, combining genetic design, nanoparticle-based imaging, immune response assays, and preclinical testing, while also applying a GAN-based model for motility prediction. Her work resulted in peer-reviewed publications and patents. She earned her M.S. in Mechanical Engineering (Bio/Nano) from Seoul National University, where she studied DNA--hydrogel composites with emphasis on mechanical resilience, responsiveness, and biocompatibility. Currently at LTYFA, she develops AI-driven genomic and biomaterial analysis pipelines, applying gene regulatory network inference to therapeutic design. Amy's expertise in complex system modeling, AI-driven analysis, and biomaterial resilience translates directly to designing and validating robust synthetic environments and adaptive materials for humanoid applications. Nijamudheen Abdulrahiman is a recognized leader in materials discovery and property prediction, with over 40 peerreviewed publications, a review article, two books, and frequent international presentations. His expertise spans semiconductors, catalysts, drug discovery, and energy materials, with strong emphasis on electronic structure calculations, molecular dynamics, and machine learning integration. As a referee for over 15 scientific journals, he brings both depth and credibility in computational modeling and HPC-based simulations. Nijam's background in atomistic simulation, ML/AI integration, and high-performance computing positions him to strengthen Omnia's ability to model material and environmental interactions in synthetic environments, improving fidelity for Armyrelevant humanoid training and survivability scenarios. Mohammed Ayman Habib specializes in intelligent robotics, embedded systems, and spatial computing for autonomy and wearables. He has hands-on experience with RADAR, LiDAR, microcontrollers, and real-time control systems for autonomous platforms, and is completing his M.S. in Electrical \& Computer Engineering at the University of Utah. His background includes designing embedded control systems for real-time perception and decision-making, integrating sensing, actuation, and autonomy pipelines. Mohammed's expertise directly supports the development of synthetic-data--trained autonomy models, real-time control for humanoid navigation, and robust embedded system integration with DoD platforms.
IV. CURRENT TECHNICAL MATURITY
The core components of Omnia's synthetic data generation pipeline (AR/VR capture, spatial browsing, physics-based sensor simulation, and automated annotation) have been validated in relevant environments. We have demonstrated transfer of perception and autonomy models trained on synthetic data into commercial robotic platforms with positive results. This places us between TRL 5 (component validation in relevant environment) and TRL 6 (prototype demonstrated in relevant environment). Our data pipeline integrates with standard AI/ML frameworks (PyTorch/TensorFlow) and exports models compatible with humanoid autonomy stacks. Interfaces for sensor simulation, mission behavior modeling, and communications have been defined and validated in the lab. This corresponds to IRL 3 (compatibility established, limited interfaces) and IRL 4 (subsystem integration validated in laboratory environment).
V. TECHNICAL RISKS
While Omnia's innovation is on track to reach TRL 7--8 by 2026, several technical risks remain on the path to a fully mature solution: 
\item  Synthetic-to-Real Transfer Fidelity: Models trained primarily on synthetic data may encounter a performance gap when exposed to unmodeled real-world conditions
(e.g., dust, weather extremes, adversary countermeasures). If not mitigated, this could limit the reliability of humanoid autonomy in operational environments. Employing hybrid training (synthetic + curated real data), adversarial scenario generation, and field validation loops to continuously calibrate synthetic environments against real sensor data. 
\item  Edge-Case Coverage in Synthetic Environments: Even high-volume synthetic datasets can miss rare or adversarial conditions (e.g., novel CBRNE signatures, unconventional urban layouts). Unaddressed edge cases could lead to brittle autonomy under battlefield stress. Expand scenario libraries through collaboration with Army SMEs; incorporate stochastic/randomized environment generation to expose autonomy to extreme edge conditions.
\item  Integration with Diverse Humanoid Hardware: While Omnia's pipeline is hardware-agnostic, integration timelines may slip if humanoid prototypes diverge in sensors, power constraints, or software stacks. Misalignment could delay demonstrating full-system capabilities in the 2026 finals. Maintaining modular APIs, conducting earlyintegration pilots with multiple platforms, and designing flexible data export layers to bridge differences.
\item  Counter-Detection \& Survivability Validation: Accurately modeling adversary EO/IR, radar, and EW systems in synthetic environments is challenging. Imperfect modeling could reduce the effectiveness of counter-detection training (HAW 007-A/B). Survivability is a core Army requirement; underestimating adversary sensors could degrade operational viability. Incorporating open-source threat intelligence and Army-provided sensor profiles into our simulation models to validate counter-detection strategies in controlled hardware-in-the-loop tests. The synthetic-to-real transfer gap is our primary concern, because it directly impacts the credibility of synthetic data as a force multiplier for humanoid development. If unmanaged, this gap could reduce trust in autonomy trained via Omnia's pipeline. That is why our technical roadmap places heavy emphasis on hybrid validation, adversarial testing, and iterative integration with Army prototypes to ensure fidelity and confidence.
VI. TECHNICAL MATURATION PLAN
Over the next eight months, Omnia will execute a structured maturation plan to address key technical risks and advance toward TRL 7--8 and IRL 7--8 readiness. In the first two months, we will expand our synthetic scenario library to capture critical edge cases, including CBRNE signatures, contested EW environments, and extreme weather or illumination, leveraging our spatial browsing pipeline and Army SME input. From months two to four, we will focus on hybrid validation of synthetic-to-real transfer by training models on mixed datasets and benchmarking them against Army-provided and commercial robotics data. Between months three and six, we will build modular APIs to ensure hardware-agnostic integration across perception, autonomy, and counter-detection modules, testing them with multiple robotics software stacks in collaboration with prototype developers. In months five through seven, we will validate survivability strategies (HAW 007-A/B) by incorporating open-source and Army-furnished adversary sensor models into our synthetic environments and running hardware-in-the-loop EO/IR experiments with sensor labs. Finally, in month eight, we will deliver an interim integrated demonstration showcasing the full pipeline---from synthetic data generation to AI model training to a humanoid prototype executing Warrior Skill Level 1 tasks such as navigation (HAW 006), perception (HAW 003/005), and autonomy behaviors (HAW 008-A/B)---partnering with a humanoid hardware vendor to prove mission-relevant capabilities in a realistic setting.
VII. PROJECTED TECHNOLOGY READINESS LEVEL
Based on our current progress, execution plan, and longterm maturation roadmap, Omnia projects reaching TRL 7--8 by July 31, 2026. Today, our system is at TRL 5--6, with validated synthetic data generation components and prototype integrations in relevant environments. Over the next 18 months, we will advance through hybrid validation, modular integration, and survivability testing to demonstrate humanoid prototypes performing Warrior Skill Level 1 tasks such as navigation, patrol, reconnaissance, and CBRNE sensing in Army-relevant field environments. By the 2026 experimentation event, we expect to complete end-to-end validation of our pipeline---synthetic data generation, AI model training, and humanoid integration---achieving measurable improvements in adaptability, survivability, and mission performance compared to traditional development. This structured progression makes TRL 7--8 both credible and achievable under self-funding.
VIII. PROJECTED INTEGRATION READINESS LEVEL
By July 31, 2026, Omnia projects reaching IRL 7--8, demonstrating that our synthetic data pipeline and autonomy modules are fully integrated into militarized humanoid prototypes and validated in operational contexts. Currently at IRL 3--4, with interfaces and subsystems validated in laboratory and simulated settings, we are progressing through modular API development, early integration pilots, and interim demonstrations. This maturation path will culminate in Army-led field trials where our autonomy, perception, and survivability modules will operate seamlessly across humanoid prototypes, tactical networks, and communications systems. By the 2026 final experimentation event, we expect to prove end-to-end interoperability---from perception to decision-making, communications, and mission execution---positioning our system to function as a reliable squad-level teammate. This trajectory aligns with the Army's Appendix B expectations (HAW 001--014) and establishes a credible pathway to full operational integration.
IX. PROJECTED TRL AND IRL CREDIBILITY
Omnia's TRL and IRL assessments are supported by proven sim-to-real transfer results, large-scale synthetic training evidence, and recent DoD validation of synthetic environments. On the TRL side, Tobin et al. (NIPS 2017) demonstrated that
models trained entirely on randomized synthetic RGB data achieved 1.5 cm accuracy in real-world grasping, proving the feasibility of synthetic-only training; Prakash et al. (CVPR 2018) showed that structured domain randomization combined with limited real data outperformed models trained solely on large real datasets such as KITTI; and Rudin et al. (Science Robotics 2022) successfully transferred locomotion policies trained entirely in simulation to ANYmal quadrupeds, achieving reliable real-world gait execution after minutes of GPU training. Together, these results substantiate our current TRL 5--6 status and support our trajectory toward TRL 7--8 by 2026, when humanoids will demonstrate validated autonomy in Army-relevant environments. On the IRL side, NVIDIA Omniverse Replicator provides a production-grade synthetic data engine with standardized APIs for perception, autonomy, and sensor modeling, enabling hardware-agnostic integration across robotics platforms and validating our current IRL 3--4. Finally, the FY2024 DOT\&E report documented the Army's STE-LTS operational demonstration, confirming that synthetic training devices improve lethality and survivability and are advancing toward rapid prototyping and fielding in FY25. This institutional validation of synthetic-first systems at IRL 7--8 maturity directly aligns with Omnia's projected IRL 7--8 by July 2026.
\end{itemize}

\end{document}